\documentclass[a4paper, 10pt, conference]{ieeeconf}      % Use this line for a4 paper

\IEEEoverridecommandlockouts                              % This command is only needed if 
\usepackage{siunitx}
\usepackage{glossaries}
\usepackage{times}
\usepackage{epsfig}
\usepackage{graphicx}
\usepackage{amsmath}
\usepackage{amssymb}
\usepackage{algorithmic}
\usepackage[font=normalsize,labelfont=bf]{caption}
\usepackage{bm} % for using \mathbb to write capitals in blackboard bold
\usepackage{subcaption}
\usepackage{eso-pic} % needed for textbox
\newacronym{PnP}{PnP}{Perspective-from-n-Points}
\newacronym{SLAM}{SLAM}{Simultaneous Localization an Mapping}
\newacronym{DNN}{DNN}{Deep Neural Network}
\newacronym{CNN}{CNN}{Convolutional Neural Network}
\newacronym{DoF}{DoF}{degrees of freedom}
\newacronym{FOV}{FOV}{field of view}
\newacronym{ITS}{ITS}{intelligent transportation systems}
\newacronym{PTP}{PTP}{Precision Time Protocol}
\newcommand{\fig}[1]{Fig.~\ref{#1}}
\newcommand{\sect}[1]{Sec.~\ref{#1}}
\newcommand{\tab}[1]{Tab.~\ref{#1}}
\newcommand{\eq}[1]{Eq.~\ref{#1}}

\newcommand{\norm}[1]{\left\lVert #1 \right\rVert}

\newcommand{\quat}{\bm{q}}
\newcommand{\quathat}{\bm{\hat{q}}}
\newcommand{\decalib}{\bm{\Phi}}
\newcommand{\decalibhat}{\bm{\hat{\Phi}}}

\newcommand{\hmat}{\bm{H}}
\newcommand{\hmathat}{\bm{\hat{H}}}
\newcommand{\rmat}{\bm{R}}
\newcommand{\pmat}{\bm{K}}
\newcommand{\phimat}{\bm{\Phi}}

\newcommand{\xvec}{\bm{x}}

\newcommand{\tvec}{\bm{t}}

\newcommand{\placetextbox}[3]{% \placetextbox{<horizontal pos>}{<vertical pos>}{<content>}
	\setbox0=\hbox{#3}% put <content> in a box
	\AddToShipoutPicture*{% add <content> to current page foreground
		\put(\LenToUnit{#1\paperwidth},\LenToUnit{#2\paperheight}){\vtop{{\null}\makebox[0pt][c]{#3}}}%
	}%
}%

\title{\LARGE \bf Targetless Rotational Auto-Calibration of Radar and Camera for Intelligent Transportation Systems}

\author{Christoph~Sch{\"o}ller*$^{12}$, Maximilian~Schnettler*$^{12}$, Annkathrin~Kr{\"a}mmer$^{12}$, Gereon~Hinz$^{12}$,\\
Maida~Bakovic$^{12}$, M{\"u}ge~G{\"u}zet$^{12}$ 
and Alois Knoll$^{2}$% <-this % stops a space
\thanks{This work has been funded by the German Federal Ministry of Transport and Digital Infrastructure as part of the project Providentia.}%
\thanks{* These authors contributed equally to this work}%
\thanks{$^{1}$fortiss GmbH, Munich, Germany}%
\thanks{$^{2}$Technical University of Munich, Munich, Germany}%
}

\begin{document}

\placetextbox{0.5}{0.96}{\fbox{
		\begin{tabular}{@{}l@{}}
		\scriptsize
		\textcopyright2019 IEEE. Personal use of this material is permitted.  Permission from IEEE must be obtained for all other uses, in any current or future media, including \\
		\scriptsize  reprinting/republishing this material for advertising or promotional purposes, creating new collective works, for resale or redistribution to servers or lists, \\
		\scriptsize  or reuse of any copyrighted component of this work in other works. 
		\end{tabular}	
}}%

\maketitle
\thispagestyle{empty}
\pagestyle{empty}

%%%%%%%%%%%%%%%%%%%%%%%%%%%%%%%%%%%%%%%%%%%%%%%%%%%%%%%%%%%%%%%%%%%%%%%%%%%%%%%%

\begin{abstract}
Most intelligent transportation systems use a combination of radar sensors and cameras for robust vehicle perception. The calibration of these heterogeneous sensor types in an automatic fashion during system operation is challenging due to differing physical measurement principles and the high sparsity of traffic radars. We propose --~to the best of our knowledge~-- the first data-driven method for automatic rotational radar-camera calibration without dedicated calibration targets. Our approach is based on a coarse and a fine convolutional neural network. We employ a boosting-inspired training algorithm, where we train the fine network on the residual error of the coarse network. Due to the unavailability of public datasets combining radar and camera measurements, we recorded our own real-world data. We demonstrate that our method is able to reach precise and robust sensor registration and show its generalization capabilities to different sensor alignments and perspectives.
\end{abstract}

%%%%%%%%%%%%%%%%%%%%%%%%%%%%%%%%%%%%%%%%%%%%%%%%%%%%%%%%%%%%%%%%%%%%%%%%%%%%%%%%

\section{Introduction}
\begin{figure}[t]
    \centerline{\includegraphics[trim=235 75 35 45,clip, width=0.93\linewidth]{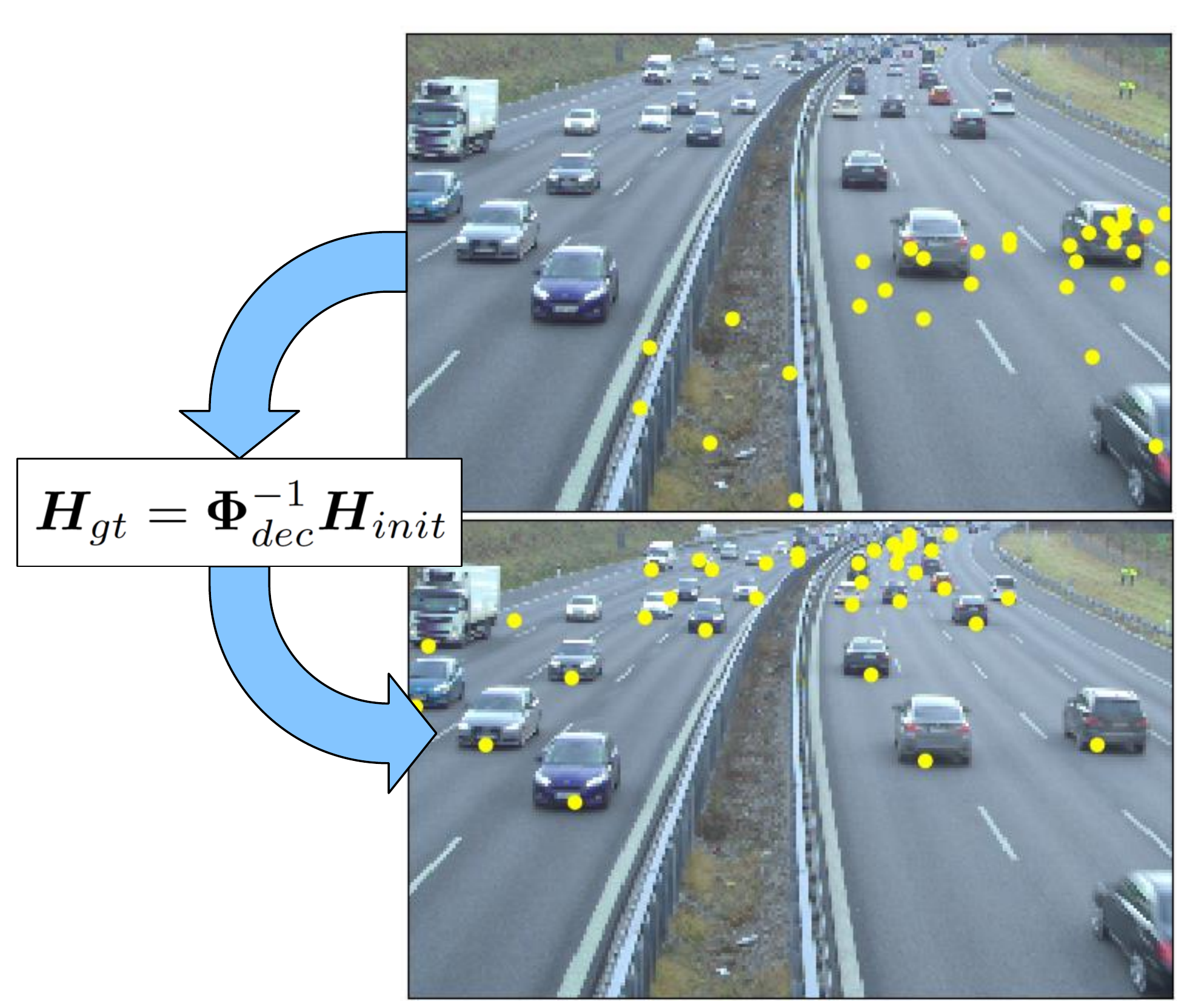}}
    \caption{The problem we solve is to estimate the correction $\decalib_{dec}^{-1}$ of the initially erroneous calibration $\hmat_{init}$ that leads to the true transformation $\hmat_{gt}$ between the radar and the camera frame. This aligns the radar's vehicle detections (yellow points) with the vehicles in the camera image.}
    \label{fig:catcher}
\end{figure}

Modern \gls{ITS} utilize many redundant sensors to obtain a robust estimate of their perceived environment. By using sensors of different modalities, the system can compensate the weaknesses of one sensor type with the strengths of another. Especially in the field of traffic surveillance and \gls{ITS}, the combination of cameras and radar sensors is common practice~\cite{Wang2017,Roy2009,Hinz2017}. Reliably fusing measurements from such sensors requires precise spatial registration and is necessary to construct a consistent environment model. A precise sensor registration can be achieved with an extrinsic calibration that results in the correct transformation between the reference frames of the sensors in relation to each other and the world. \fig{fig:catcher} demonstrates the effects of an accurate sensor calibration. The upper image shows uncalibrated sensors, where the projected radar detections do not align with the vehicles. After the extrinsic calibration each detection overlays with its corresponding object in the image.

Manual sensor calibration is a tedious and expensive process. Especially in multi-sensor systems automatic calibration is crucial to handle the growing number of redundant sensors. Here manual calibration does not scale. Additionally, this technique is infeasible for automatic online recalibration, which is necessary to account for changes to the sensor system. These decalibrations occur frequently in real world applications, for example due to vibrations, wear and tear of the sensor mounting, or changing weather conditions. Furthermore, these calibration methods should be independent of explicitly provided calibration targets, as their installation into such systems or their observed scenes is impractical and would suffer from deterioration as well. 

Calibrating systems with radar sensors and cameras is challenging due to different physical measurement principles and the high sparsity of radar detections. For the calibration without specific targets, a complex association problem between the sensors' measurements must be solved. As traffic radars do not provide visual features, such as edges, corners or color to easily associate detections with vehicles in the camera image, this association problem must be solved solely based on the relative spatial alignment and estimated distance measures between the vehicles.

In this paper we present --~to the best of our knowledge~-- the first method for the automatic calibration of radar and camera sensors without explicit calibration targets. We focus on the rotational calibration between sensors because of its high influence on the spatial registration between cameras and radar sensors in \gls{ITS}, especially for large observation distances. On the other hand, the projective error caused by translational miscalibrations in the centimeter range is negligible and easy to minimize in static scenarios by measuring with modern laser distance meters. To solve the problem of rotational auto-calibration, we propose a two-stream \gls{CNN} that is trainable in an end-to-end fashion. We employ a boosting-inspired training algorithm, where we first train a coarse model and afterwards transform the training data with its correction estimates to train a fine model on the residual rotational errors. We evaluate our approach on real-world data, recorded on the German highway A9 and show that it is able to achieve precise sensor calibration. Furthermore, we demonstrate the generalization capability of our approach by applying it to a previously unobserved perspective.

\section{Related Work}
Much research has been done on calibrating multi-sensor systems with homogeneous sensors (e.g., camera to camera), resulting in various state-of-the-art target-based and targetless calibration methods. However, it is a challenging problem to calibrate heterogeneous sensors with different physical measurement principles. While camera images provide dense data in form of pixels, lidar and even more so traffic radar sensors only record sparse depth data without color information. In this case it is difficult to match corresponding features between the sensors' measurements for calibration.

Classic approaches for the calibration of camera and laser-based depth sensors use planar checkerboards as dedicated calibration targets \cite{Zhang2004,Zhi2007}. These techniques achieve very precise estimates of the relative sensor poses, but require prepared calibration scenes. Approaches without physical targets calibrate the sensors by matching features of the natural scenes. In manual methods, a human has to pair the corresponding features in the image and depth data by hand~\cite{Scaramuzza2007}. For automatic classic approaches the capabilities regarding decalibration ranges and parameter extraction are limited~\cite{Chien2016,Levinson2013}. These drawbacks restrict their application scope and prevent them from being used for automatic calibration during system operation, which is possible with our method.

Recently, the task of sensor calibration has been approached using deep learning techniques. Early applications of \glspl{CNN} in this field focus on camera relocalization~\cite{Kendall2015}. With RegNet, Schneider et al.\@~\cite{RegNet} presented the first \gls{CNN} for camera and lidar registration, which performs feature extraction, matching, and pose regression in an end-to-end fashion on an automotive sensor setup. Their method is able to estimate the extrinsic parameters and to compensate decalibrations online during operation. To refine their result the authors use multiple networks, trained on datasets with different calibration margins. In contrast, we do not need to define calibration margins for our networks, as our second network specializes on the errors of the first network by design. Liu et al.\@~\cite{Liu2018} apply this method to the calibration of three sensors by first fusing a depth camera and a lidar that were calibrated with RegNet, and then they use the resulting dense point cloud for the calibration to a camera. Iyer et al.~\cite{CalibNet} propose CalibNet, which they train with a geometric and photometric consistency loss of the input images and point clouds, rather than the explicit calibration parameters. Due to difficulties in estimating translation parameters in a single run, they first estimate the rotation and use it to correct the depth map alignment. Then they feed the corrected depth map back into the network to predict the correct translation. However, in contrast to our approach these methods use lidar sensors with relatively dense point clouds compared to the measurements of traffic radars.

The measurement characteristics of radars cause a lack of targetless calibration methods. Traffic radars output preprocessed measurement data in form of detected objects. They lack descriptive visual features and are sparse. Additionally, measurement noise,  missing object detections, and false positives make the calibration of radars with sensors of different modality particularly challenging. Existing approaches for the calibration of multi-sensor systems with radars rely on dedicated targets, such as corner reflectors or plates, based on conductive material that ensures reliable radar detections~\cite{Helmick1993,Li2006}. Recently, these calibration concepts were extended towards the combination of radars with other sensor types. Especially the calibration with cameras is challenging, as the sensors do not share common features such as color, shapes or depth. Natour et al.~\cite{Natour2015} calibrate a radar-camera setup by optimizing a non-linear criterion, obtained from a single measurement with multiple targets and known inter-target distances. However, the targets in the radar and image data are extracted and matched manually. Per{\v{s}}i{\'{c}} et al.\@~\cite{Persic2017} designed a triangular target to calibrate a 3D lidar and an automotive radar. They experienced variable error margins in the estimated calibrations due to the sparse and noisy radar data and the geometric properties of their sensor setup. As a result, an additional optimization step using a priori knowledge of the specified radar field-of-view refines these estimated parameters. Song et al.~\cite{Song2017} use a radar-detectable augmented reality marker for a traffic surveillance system based on a 2D radar and camera, enabling an analytic solution of the paired measurements. However, there is a lack of approaches for automatic and targetless radar-camera calibration which we address in this work.

\section{Problem Statement}
\label{sec:problem-statement}

To calibrate a radar and camera to each other, the transformation that correctly projects the radar detections into the camera image must be estimated. This is the case when each projected detection spatially aligns with its corresponding object in the image. As we use a traffic radar, the detected objects are vehicles as shown in~\fig{fig:catcher}. However, our approach is not limited to the traffic domain.

The described projection of detections into the image can be computed by
\begin{equation} 
    z_c\left[\begin{array}{c}u\\v\\ 1\end{array}\right]=\pmat\hmat\xvec,
    \label{eq:projection}
\end{equation}
where $\xvec\in\mathbb{R}^{3}$ is the position of a detected vehicle in the radar coordinate system, $[u,v]^T$ are its corresponding pixel coordinates and $z_c$ the straight-line distance of the detected vehicle to the image plane of the camera, i.e.\@ the depth of the projected pixel. The projection matrix $\pmat\in\mathbb{R}^{3\times 4}$ is based on the intrinsic camera parameters and $\hmat\in\mathbb{R}^{4\times 4}$ is the extrinsic calibration matrix. The latter represents the camera pose relative to the radar and is defined as 
\begin{equation}
    \hmat = \left[\begin{array}{cc}
         \rmat & \tvec  \\
         \bm{0}^T & 1
    \end{array}\right],
\end{equation}
with $\rmat\in SO(3)$ being the rotational and $\tvec\in\mathbb{R}^{3}$ the translational component. While $\pmat$ can be estimated in a controlled calibration setting prior to deploying the sensors, $\hmat$ must be determined after deployment in situ. In our work we focus on computing the rotational component as -- compared to the translational component -- it is hard to measure and has a high impact on the quality of the inter-sensor registration, especially for large observation distances.

Our goal is to estimate the transformation $\hmat_{gt}$, that describes the true relative pose between the two sensors. A correct estimate results in the alignment of projected radar measurements and vehicles in the image. Assuming an initially incorrect calibration $\hmat_{init}$, we need to determine the present decalibration transformation $\phimat_{dec}$ that represents the error between $\hmat_{init}$ and $\hmat_{gt}$ and thus
\begin{equation} \label{eq:relation_hgt_hinit}
    \hmat_{init}=\phimat_{dec}\hmat_{gt}.
\end{equation}
In fact, we directly estimate $\phimat_{dec}^{-1}$, since it can be used to recover the correct calibration $\hmat_{gt}$ without additionally inverting $\phimat_{dec}$.

\section{Our Approach}
Our objective is to regress the relative orientation of a camera with respect to a radar sensor. To achieve this, an association problem between the radar detections and the vehicles in the camera image must be solved. This is a difficult problem, as radar detections do not contain descriptive features. A neural network can learn how to solve this association problem based on the spatial alignment between the projected radar detections and the vehicles in the image.

Our approach leverages two convolutional neural networks, where we train the first coarse network on the initially decalibrated data and then a fine network on its residual error. Both models share the same architecture, loss and hyperparameters. In this section we first explain the model and then the training process in detail.

\subsection{Architecture}
\label{sec:architecture}
\begin{figure*}
    \centerline{\includegraphics[trim=26 54 20 62,clip,width=0.85\textwidth]{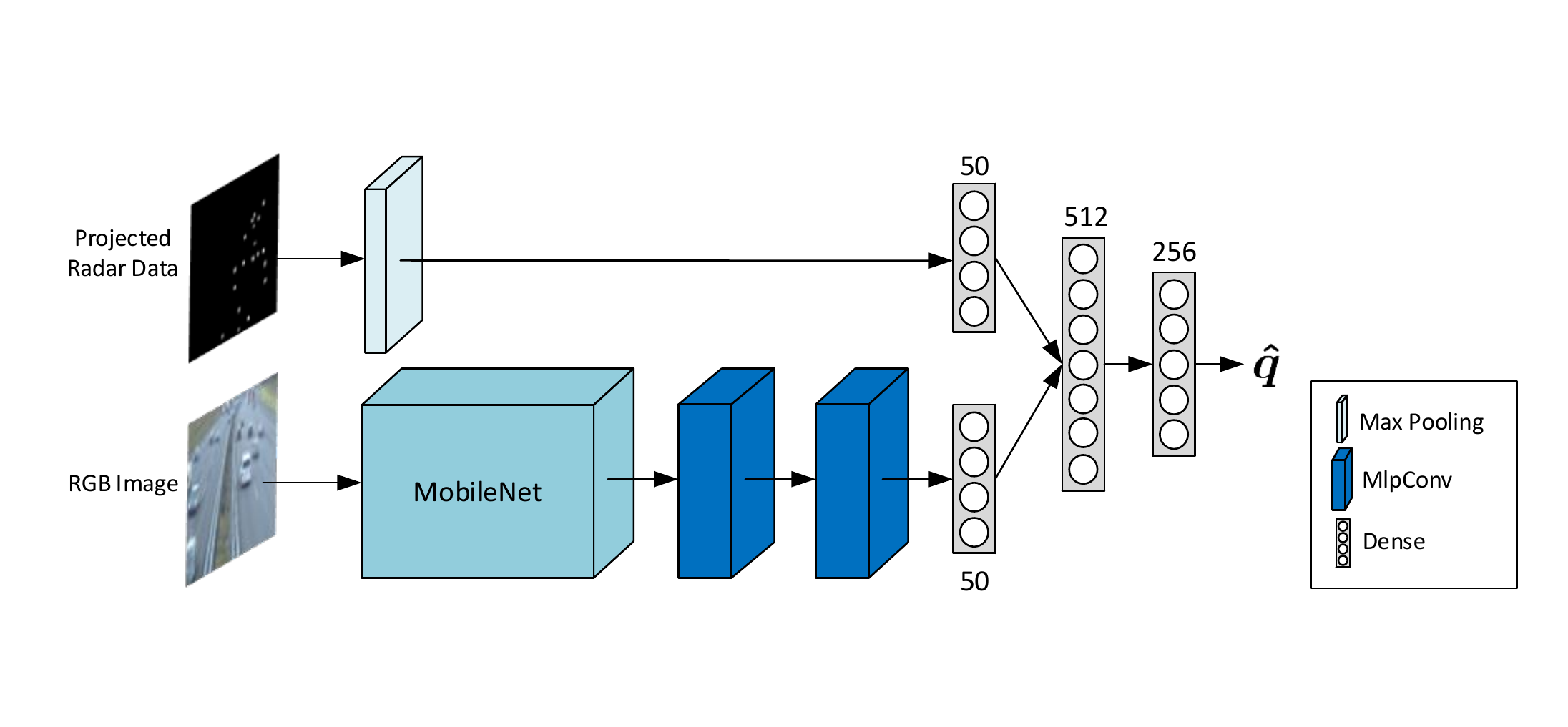}}
    \caption{The architecture of our network consists of two input streams, one for radar projections and one for rgb images. Both streams end in a 50 neuron fully-connected layer to condense information. Then we regress for an output quaternion $\quathat$ that describes the calibration correction.}
    \label{fig:architecture}
\end{figure*}

Our model is built as a two-stream neural network, consisting of a rgb-input and a radar-input as shown in~\fig{fig:architecture}. It outputs a transformation to correct the rotational error of the calibration between respective camera and radar as a quaternion.

The rgb-input is a camera image and the radar-input is a sparse matrix with radar projections. The image is standardized and resized to a resolution of $240\times150$ pixels. It gets propagated through the rgb stream of our network that starts with a cropped MobileNet~\cite{Howard2017} with width multiplier 1.0. We crop the MobileNet after the third depthwise convolution layer (conv\_dw\_3\_relu) to extract low-level features, while preserving spatial information. We use a MobileNet that has been pre-trained on ImageNet~\cite{JiaDeng2009}, but include the layers for fine-tuning in further training. The MobileNet is followed by two MlpConv~\cite{Lin2013} layers, each consisting of a 2D convolution with kernel size $5\times5$, followed by two $1\times1$ convolutions and 16 filter maps in each component. The task of the rgb stream is to detect vehicles and to estimate where radar detections will occur.

The radar-stream receives the projected radar detections with the same resolution as the camera image as input. Each projection occupies one cell in the sparse matrix and stores the inverse depth $1/z_c$ of respective projected detection, as proposed by~\cite{RegNet}. We apply a $2\times2$ max-pooling to reduce the input dimension to a feasible size and do not use convolutions in the radar stream to retain the sparse information.

Then we embed each stream into a 50 dimensional latent vector using a fully-connected layer. This latent vector contains the input information in a dense and compressed format. The following regression block consists of three layers with 512, 256 and 4 neurons. Between the first two layers we apply dropout regularization~\cite{Dropout2014}. The four output neurons correspond to the components of the quaternion that describes the calibration correction. We use linear activations for the final output layer, and PReLu~\cite{PReLu2015} activations everywhere else, except in the MobileNet block. This empirically lead to better performance compared to classic ReLu activations. The task of the regression block is to estimate the rotational correction that solves the misalignment between the camera image and the radar detections.

\subsection{Loss Function}
We use the Euclidean distance as the loss function between the true quaternion $\quat$, and the predicted quaternion $\quathat$ that represents the estimated correction of the decalibration. 

The Euclidean distance is a common distance measure to define a rotational loss function over quaternions~\cite{RegNet,Kendall2015}. Since this metric is ambiguous and can lead to different errors for the same rotations~\cite{Huynh2009}, we also evaluated the performance of our approach using the geodesic quaternion metric
 \begin{equation}
    \label{eq:quatmetric}
    \mathcal{L}_{\theta} = 1 - |\quat \cdot \frac{\quathat}{\norm{\quathat}}| + \alpha |1 - \norm{\quathat}|,
 \end{equation}
proposed by \cite{Kuffner2004}. We added a length error term, weighted by $\alpha$ that we empirically evaluated to $0.005$. Without this additional length term the network's output diverges and the learning plateaus. As this loss resulted in similar performance despite its theoretical superiority, we finally used the Euclidean distance $\norm{\quat - \quathat}$ to save an additional hyperparameter.

\subsection{Hyperparameters}
\label{sec:params}
We use the Adam optimizer~\cite{KingmaB14} with the parameters proposed by its authors and learning rate $0.002$, that we reduce by a factor of $0.2$ once the validation loss plateaus for five epochs. To initialize our weights we use orthogonal initialization~\cite{Saxe2014}. For the dropout we set a probability of $0.5$, use a batch size of 16 and early stopping when the validation loss does not improve for 10 epochs.

\subsection{Cascaded Residual Learning}
\label{sec:learning-algorithm}
To improve the calibration results of our first, coarse network that we train on the original data, we train a second, fine network on the remaining residual error. This is inspired by gradient boosting algorithms~\cite{friedman2002stochastic}, where each subsequent learner is trained on the residual error of the previous one.

This boosting has multiple advantages that lead to more accurate calibration parameters. During operation, the first network roughly corrects the initial calibration error and the sensors are approximately aligned. For the second network more radar detections can be projected into the camera image, which leads to a higher number of correspondences that enable the second network to perform a more fine-grained correction. Furthermore, the second network implicitly focuses on solving the errors in those axes that the first network performed poorly on. In our case, the fine network performs much better on solving the roll error, as errors around the $z$-axis of the camera cause only relatively small projective discrepancy, and thus the coarse network focuses on tilt and pan.

In detail, we train the first network on dataset $D$, for which the radar detections of each sample were projected with a transformation $\hmat_{init, i}$, obtained by applying a random decalibration $\decalib_{dec,i}$ to the true calibration $\hmat_{gt}$. Index $i$ refers to the sampled decalibration. After the first network's training we transform $D$ into a new dataset $D'$, on which we train the second network. $D'$ contains training samples corrected by the output of the first network. We transform the samples by converting the output quaternion $\quathat_i$ for each sample to transformation $\decalibhat^{-1}_{dec,i}$. Then we compute a new, corrected extrinsic matrix
\begin{equation}
\hmathat_{gt, i} = \decalibhat^{-1}_{dec,i} \hmat_{init, i}
\end{equation}
for each sample and reproject the corresponding radar detections as described by \eq{eq:projection}. We obtain the correction of the residual decalibration by
\begin{equation}
 \decalib'^{-1}_{dec,i} = \decalib^{-1}_{dec,i} \decalibhat_{dec,i},
\end{equation}
which serves as the new label in the transformed dataset $D'$. The second network is then trained on $D'$.

At inference time we obtain the approximate true calibration for a new sample by computing
\begin{equation}
 \hmathat_{gt} = \decalibhat'^{-1}_{dec} \decalibhat^{-1}_{dec} \hmat_{init},
\end{equation}
where $\decalibhat'^{-1}_{dec}$ is the output of the second network and $\decalibhat^{-1}_{dec}$ of the first network. Note that before computing $\decalibhat'^{-1}_{dec}$ we perform a reprojection in the same way as during training.

\subsection{Iterative and Temporal Refinement}
\label{sec:refinement}
In the field of \gls{ITS} and autonomous driving, sensor data is usually available as a continuous stream. A single decalibration of the sensor setup is more likely than completely random decalibrations for each sample. A temporal average over correction estimates for multiple consecutive samples can reduce the influence of estimation errors made for individual samples and thus increase the robustness and accuracy of our method.

\section{Experiments}
In this section we explain which data we used to train and evaluate our approach. Furthermore, we explain our evaluation process in detail and present quantitative, as well as qualitative results.

\subsection{Dataset}
\label{sec:dataset}

\begin{figure}[t]
    \centerline{\includegraphics[trim=0 30 0 0,clip,width=0.80\linewidth]{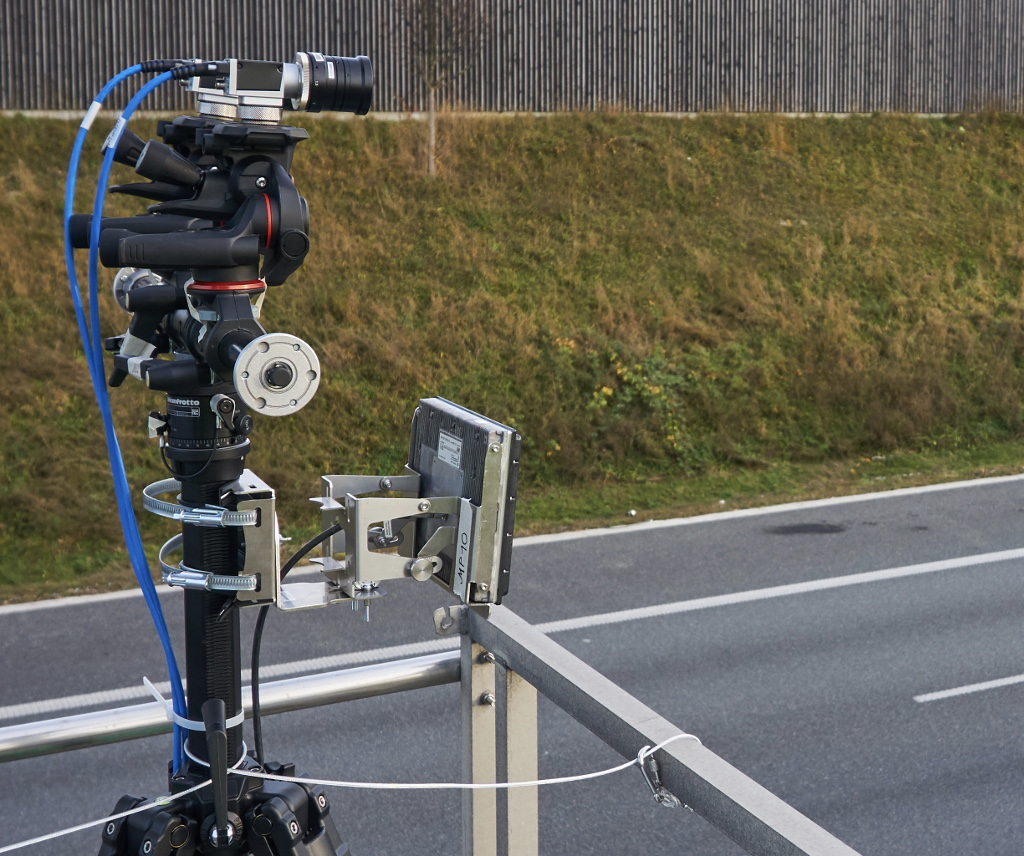}}
    \caption{Our sensor setup with a radar and camera above the highway.}
    \label{fig:sensor_setup}
\end{figure}

In the field of \gls{ITS}, public datasets containing data of radars combined with cameras are not available. Therefore, we generated our own dataset using sensor setups developed within the scope of the research project Providentia~\cite{Hinz2017}.
Two identical setups were installed on existing gantry bridges along the Autobahn~A9, overlooking a total of eight traffic lanes. Our sensor setup is shown in~\fig{fig:sensor_setup} and consists of a Basler \mbox{acA1920-50gc} camera with a lens of \SI{25}{\milli\metre} focal length and a smartmicro \mbox{UMRR-0C} Type~40 traffic radar.

The camera records rgb images with a resolution of ${1920\times1200}$ pixels, while the radar outputs vehicle detections as positions. The radar measurements can result in undetected vehicles, multi-detections for large vehicles like trucks or buses, and false positives due to measurement noise.

\subsubsection{Ground Truth Calibration}
Since our approach requires a reference transformation $\hmat_{gt}$ between the sensors, we put special effort and care on the initial manual calibration. This is equivalent with manual labeling in other supervised learning problems.

We calibrated the cameras intrinsically with a checkerboard based method in our laboratory, while the radar is intrinsically calibrated ex-factory. The translational extrinsic parameters of the sensor setup were manually measured on-site with a spirit level and laser distance meter. We estimated the initial rotation parameters of the sensors with respect to the road using vanishing point based camera calibration~\cite{Kanhere2010} (one vanishing point, known height above the road and known focal length) and the internal calibration function of the radar sensor. Afterwards, we fine-tuned the extrinsic rotational parameters by minimizing the visual projective error.

\subsubsection{Training Data Generation}
To obtain the necessary number of samples to train and evaluate our networks, it would be infeasible to record with many different sensor setups and manually determine each groundtruth calibration. Therefore, we randomly distorted the calibration $\hmat_{gt}$ for one sensor setup per measurement point as proposed by \cite{RegNet}. In particular, we randomly generated 6-\acrshort{DoF} decalibrations $\phimat_{dec}$ for each sample and used these decalibrations to compute initial decalibrated extrinsic matrices $\hmat_{init}$, according to~\eq{eq:relation_hgt_hinit}. Afterwards, we projected the radar detections on the image according to~\eq{eq:projection}, leading to a mismatch between the detections and the vehicles in the image. Besides, we filtered generated samples with less than 10 remaining correspondences. This ensures the exclusion of training samples without correspondences between camera and radar projection, with which learning is not possible.

In particular, the decalibration angles were sampled from a uniform distribution on $[\SI{-10}{\degree}, \SI{10}{\degree}]$ for the tilt and pan, and $[\SI{-5}{\degree}, \SI{5}{\degree}]$ for the roll angle. We assumed a smaller roll decalibration as this angle is easier to measure with a spirit level. We multiplied resulting matrices into a single rotational decalibration. Furthermore, we added a translation error with a standard deviation of $\SI{10}{\centi\meter}$. Even though translation errors are minimal as distances are easy to measure, by this we account for errors during the manual calibration process and show that our approach is robust to it. Creating our dataset as described resulted in a total of $37929$ samples for the first sensor setup, of which we used $34137$ for training and $3792$ for validation. Additionally, we generated an independent test set $\mathcal{T}_1$ with $2536$ samples, where we only included images and radar detections that do not appear in the training data. We further generated a second test set $\mathcal{T}_2$ with $2012$ samples from a different sensor setup on a second gantry bridge in the same manner to evaluate the generalization of our approach.

\subsection{Evaluation}

We trained our models with the boosting-inspired approach described in~\sect{sec:learning-algorithm} and the dataset generated with random decalibrations as explained in~\sect{sec:dataset}.

\subsubsection{Random Decalibration}

\begin{table}[b]
\centering
\begin{tabular}{lllll}
\hline
               & \multicolumn{1}{c}{Tilt} & \multicolumn{1}{c}{Pan} & \multicolumn{1}{c}{Roll} & \multicolumn{1}{c}{Total} \\ \hline
Initial        & \SI{4.45}{\degree}             & \SI{4.95}{\degree}            & \SI{2.52}{\degree}             & \SI{7.90}{\degree}              \\
Coarse Network & \SI{0.46}{\degree}             & \SI{0.81}{\degree}           & \SI{2.43}{\degree}             & \SI{2.78}{\degree}              \\
Fine Network   & \textbf{0.21$^\circ$}          & \textbf{0.32$^\circ$}         & \textbf{1.32$^\circ$}          & \textbf{1.45$^\circ$}           \\ \hline
\end{tabular}
\caption{Mean absolute errors for all axes and in total initially, after applying the coarse network and after applying the fine network on the test set $\mathcal{T}_1$. The fine model focuses on correcting the remaining roll error and significantly improves tilt and pan as well.}
\label{tab:validation-results}
\end{table}

\begin{figure*}[ht]
    \centering
    \begin{subfigure}[b]{0.5\textwidth}
        \includegraphics[trim=0 5 0 77,clip,width=0.99\textwidth]{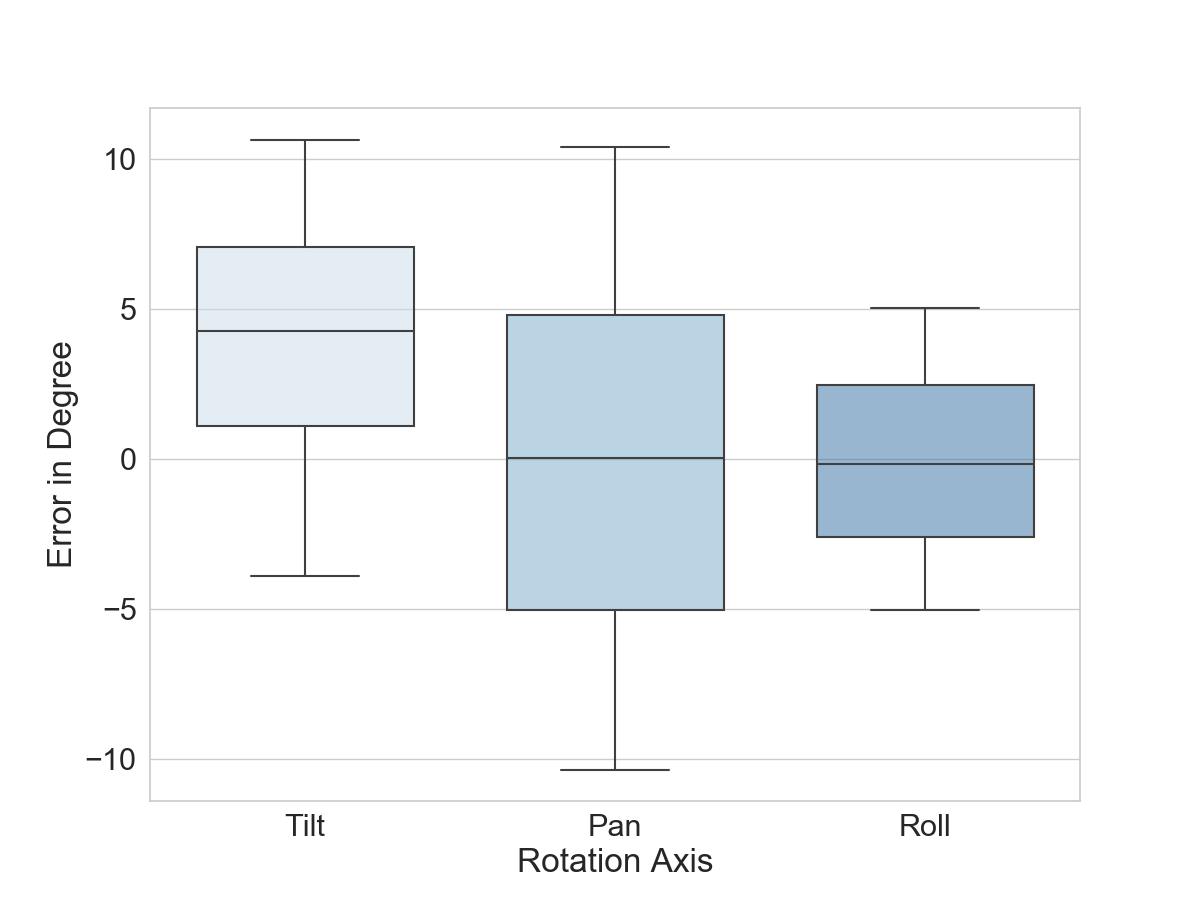}
        \caption{Initial}
        \label{fig:error-distributions-a}
    \end{subfigure}%
    \begin{subfigure}[b]{0.5\textwidth}
        \includegraphics[trim=0 5 0 77,clip,width=0.99\textwidth]{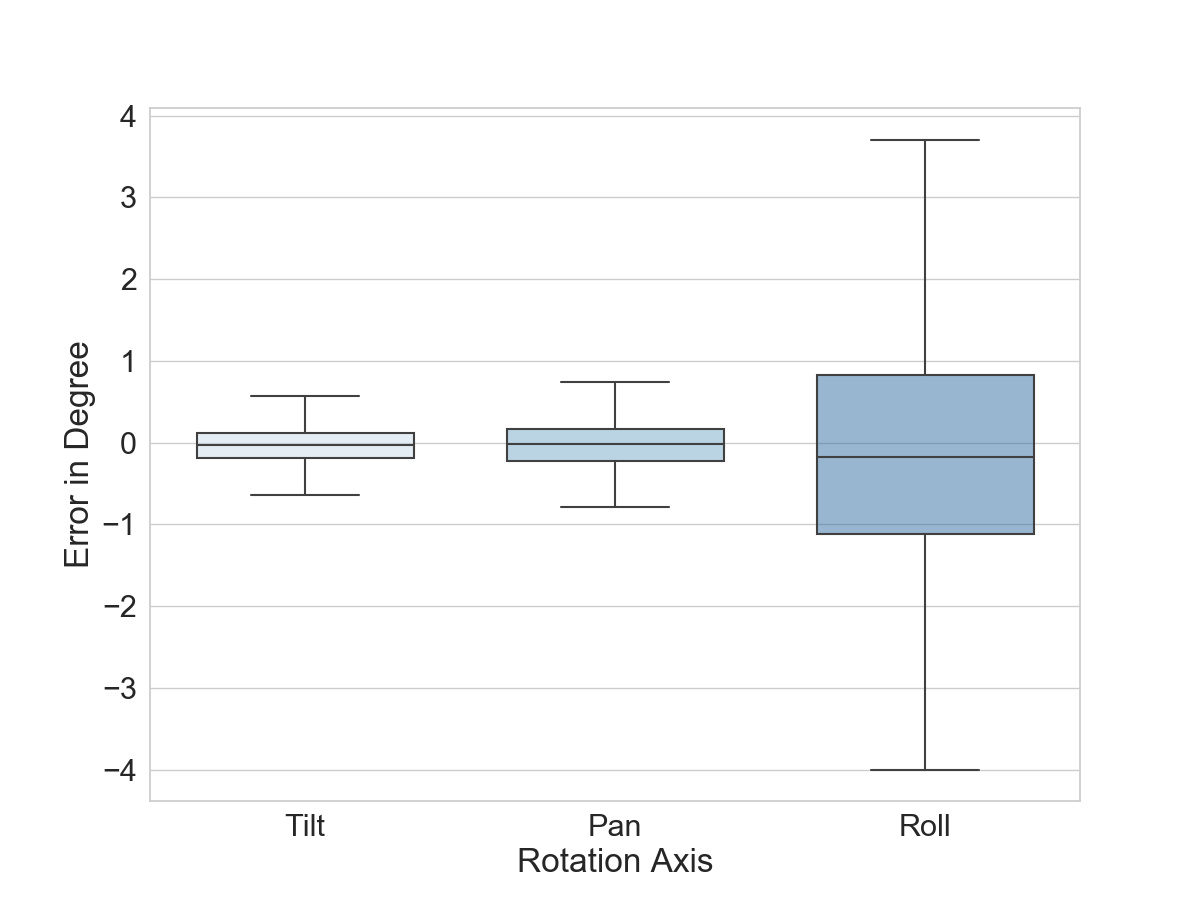}
        \caption{Calibration Result}
        \label{fig:error-distributions-b}
    \end{subfigure}%
    \caption{Initial and resulting errors for each rotation angle on test set $\mathcal{T}_1$ with random calibration errors. Note that the positive shift for the tilt angle errors is a result of filtering the samples with less than 10 projected detections in the image, as described in~\sect{sec:dataset}. Tilting the camera downwards likely moves the detections out of the upper image border.}
    \label{fig:error-distributions}
\end{figure*}

\begin{figure*}[ht]
    \centering
    \par\medskip
    \begin{subfigure}[b]{0.33\textwidth}
        \includegraphics[trim=53 55 46 60,clip,width=0.988\textwidth]{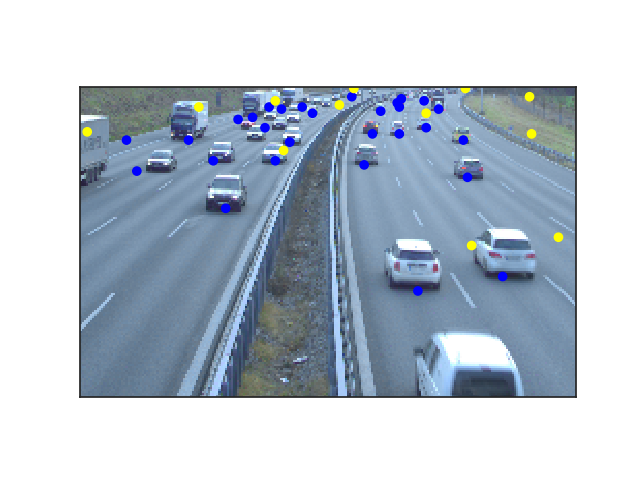}
    \end{subfigure}%
    \begin{subfigure}[b]{0.33\textwidth}
        \includegraphics[trim=53 55 46 60,clip,width=0.988\textwidth]{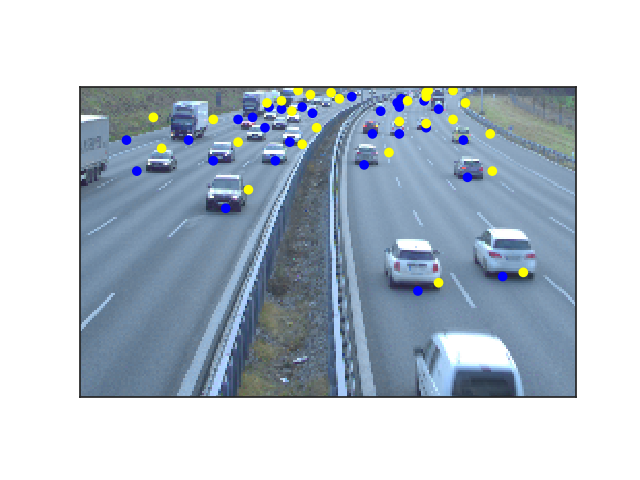}
    \end{subfigure}%
    \begin{subfigure}[b]{0.33\textwidth}
        \includegraphics[trim=53 55 46 60,clip,width=0.988\textwidth]{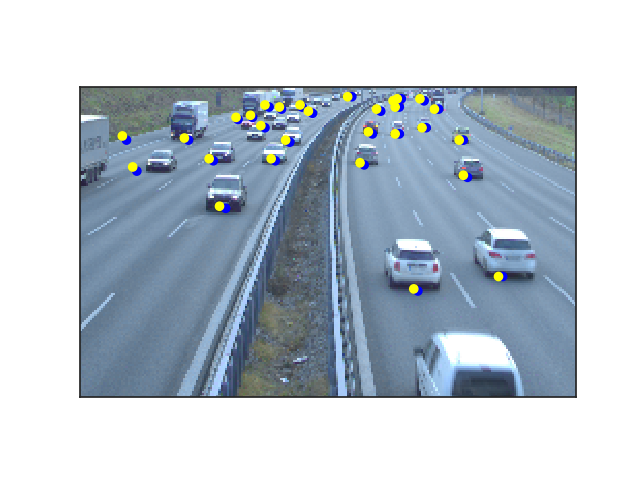}
    \end{subfigure}%
    \par\medskip
    \begin{subfigure}[b]{0.33\textwidth}
        \includegraphics[trim=53 55 46 60,clip,width=0.988\textwidth]{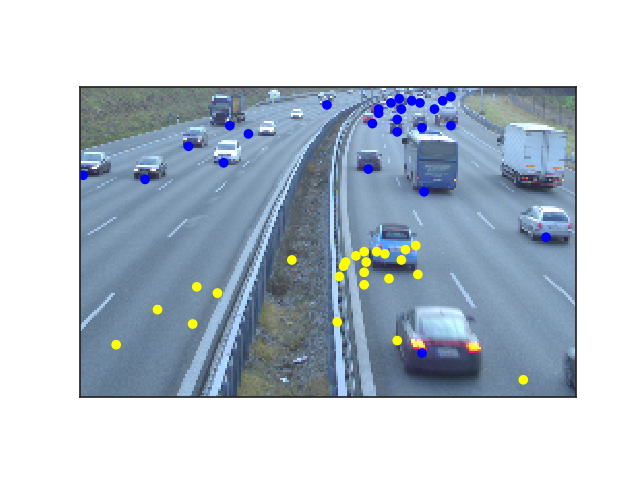}
        \caption{Initial}
        \label{fig:qualitative-examples-initial}
    \end{subfigure}%
    \begin{subfigure}[b]{0.33\textwidth}
        \includegraphics[trim=53 55 46 60,clip,width=0.988\textwidth]{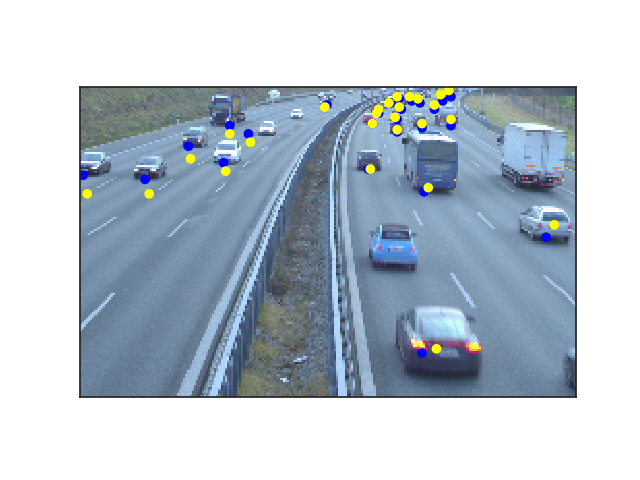}
        \caption{Coarse Network}
        \label{fig:qualitative-examples-coarse}
    \end{subfigure}%
    \begin{subfigure}[b]{0.33\textwidth}
        \includegraphics[trim=53 55 46 60,clip,width=0.988\textwidth]{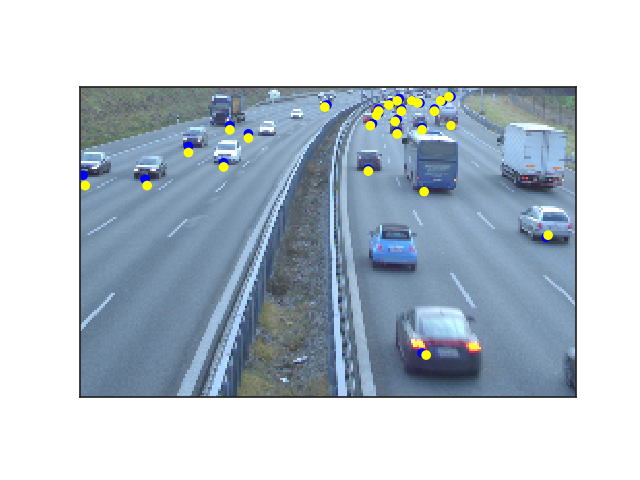}
        \caption{Fine Network}
        \label{fig:qualitative-examples-fine}
    \end{subfigure}%
    \caption{Each row depicts the application of our model to a decalibrated sample from test set $\mathcal{T}_1$. Blue points represent the projected radar detections using the ground truth calibration and yellow projections using the calibration at (a) the initial stage, (b) after applying the coarse network and (c) after applying the fine network. While the coarse network achieves a reasonable, but still imprecise calibration, the fine model handles the precise adjustment.}
    \label{fig:qualitative-examples}
\end{figure*}

\tab{tab:validation-results} shows the average angular errors of our networks using test set $\mathcal{T}_1$ with random decalibrations. While the coarse network achieves significant improvements in the tilt and pan angles, it struggles to correct the roll. The roll calibration error is weaker correlated with the input as it has only a small projective influence over the long distances we work with. However, our fine model decreases the roll error significantly as it has more influence on the total remaining projective discrepancy after the coarse correction step. In total, we achieve a mean error reduction of \SI{95.3}{\percent} in tilt, \SI{93.5}{\percent} in pan and \SI{47.7}{\percent} in roll over the initial decalibrations. The remaining errors after our correction are approximately normally distributed around zero, which means our approach works reliably with only few outliers and can be trusted in a real-world setting~(\fig{fig:error-distributions}).

In \fig{fig:qualitative-examples} we demonstrate qualitative examples of applying our approach to different decalibration scenarios. The main task of the coarse network is to find the right correspondences between radar detections and vehicles in the images. Based on these correspondences it estimates a rough correction for the initial decalibration. In case of decalibrations with only few successfully projected detections, the network's correction leads to more projections onto the image plane that are then provided to the fine network. This effect can be observed in the first row in \fig{fig:qualitative-examples}. The fine network makes use of the increased number of correspondences and refines the calibration as shown in column \textbf{(c)}. It is particularly good at correcting rotational errors in the roll direction. In the second row \textbf{(b)} it can be observed that the coarse network is not able to solve the roll error because it has a relatively small impact on the projection discrepancy. The yellow points in the left half of the image are rotated below, and in the right half of the image above the blue ground truth detections. The fine network in the second row \textbf{(c)} was able to correct this residual error.

\subsubsection{Static Decalibration}

\begin{figure*}[ht]
    \centering
    \begin{subfigure}[b]{0.5\textwidth}
        \includegraphics[trim=0 5 0 77,clip,width=0.99\textwidth]{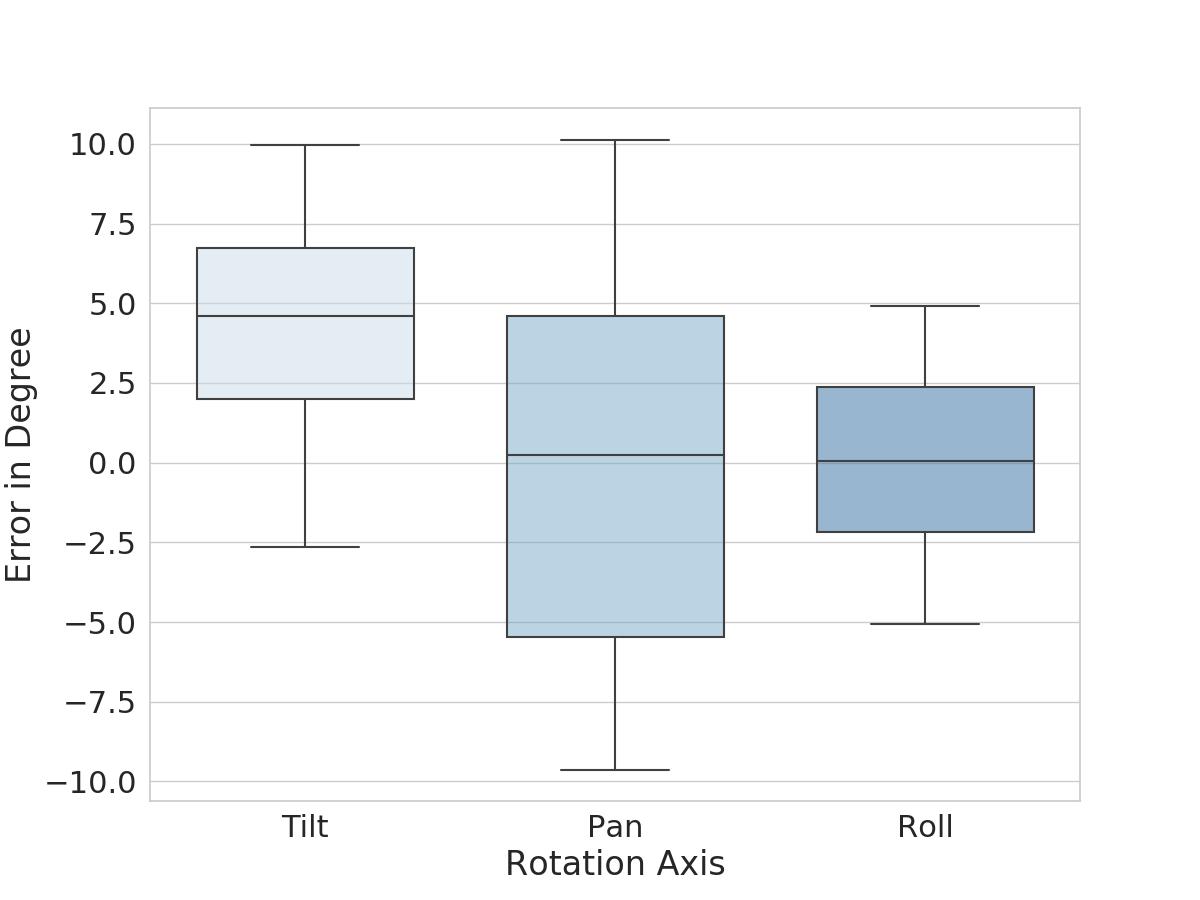}
        \caption{Initial}
    \end{subfigure}%
    \begin{subfigure}[b]{0.5\textwidth}
        \includegraphics[trim=0 5 0 77,clip,width=0.99\textwidth]{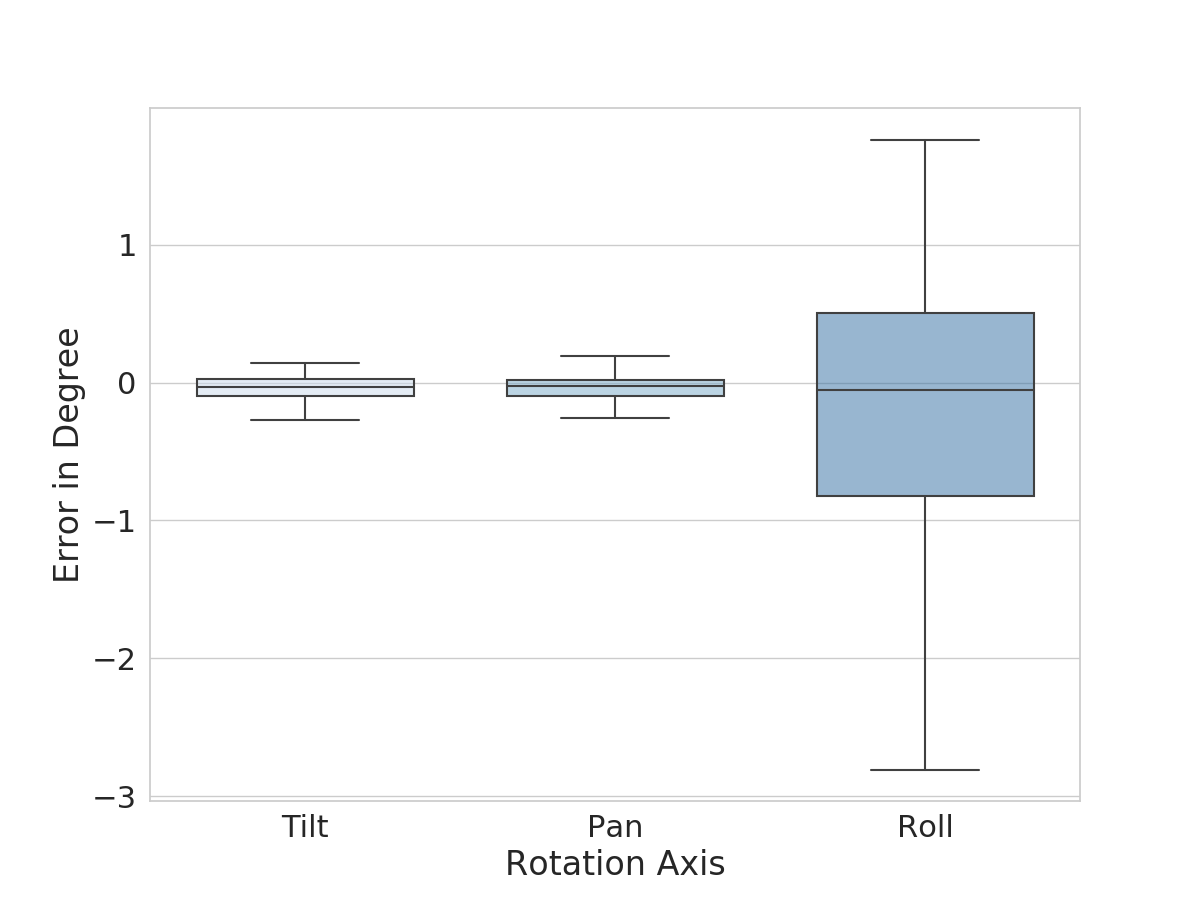}
        \caption{Result}
    \end{subfigure}%
    \caption{Initial and resulting distributions of the errors for the 100 static decalibrations for test set $\mathcal{T}_1$. For each decalibration we averaged all sample errors.}
    \label{fig:error-distributions-static-all}
\end{figure*}

\begin{figure*}[ht]
    \centering
    \begin{subfigure}[b]{0.5\textwidth}
        \includegraphics[trim=0 5 0 77,clip,width=0.99\textwidth]{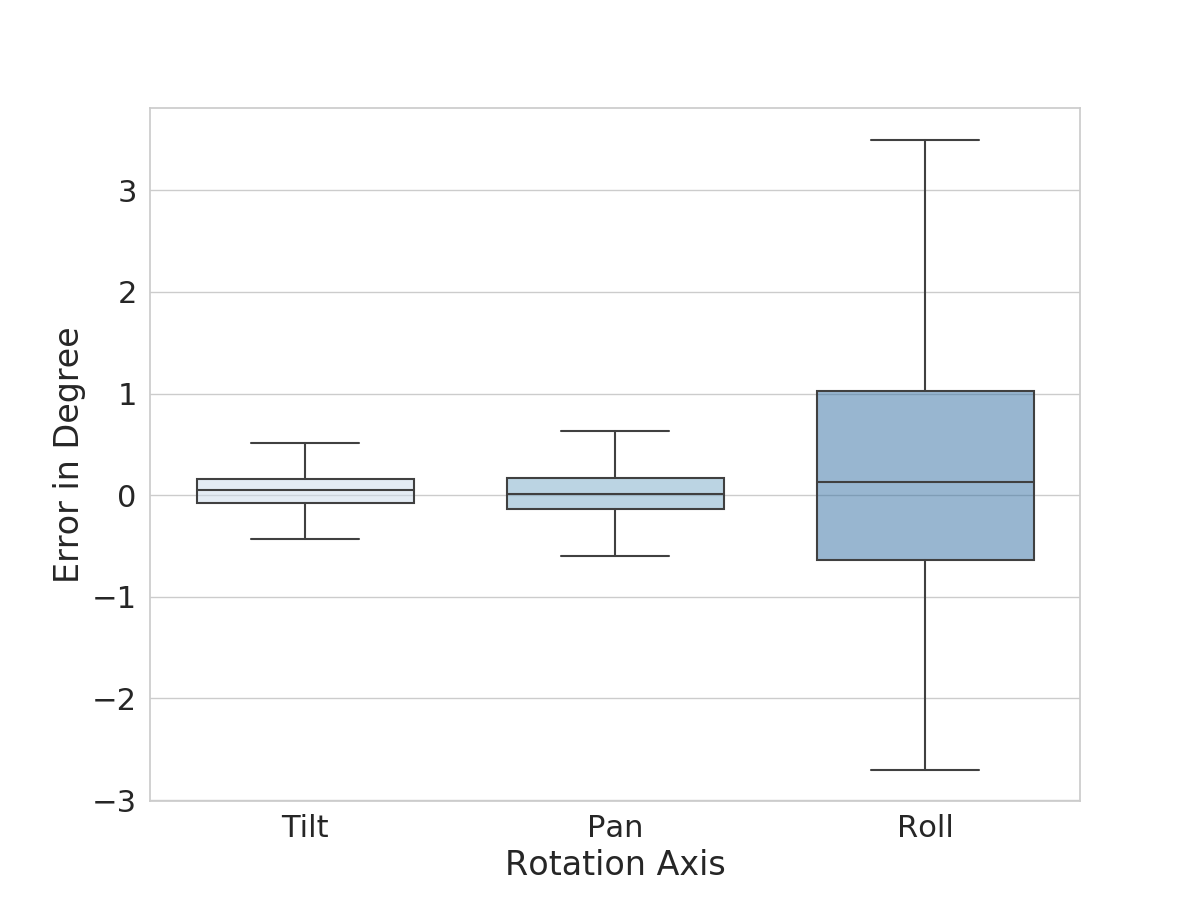}
        \caption{Initial errors: tilt = \SI{3.21}{\degree}, pan = \SI{6.24}{\degree}, roll = \SI{1.75}{\degree}}
    \end{subfigure}%
    \begin{subfigure}[b]{0.5\textwidth}
        \includegraphics[trim=0 5 0 77,clip,width=0.99\textwidth]{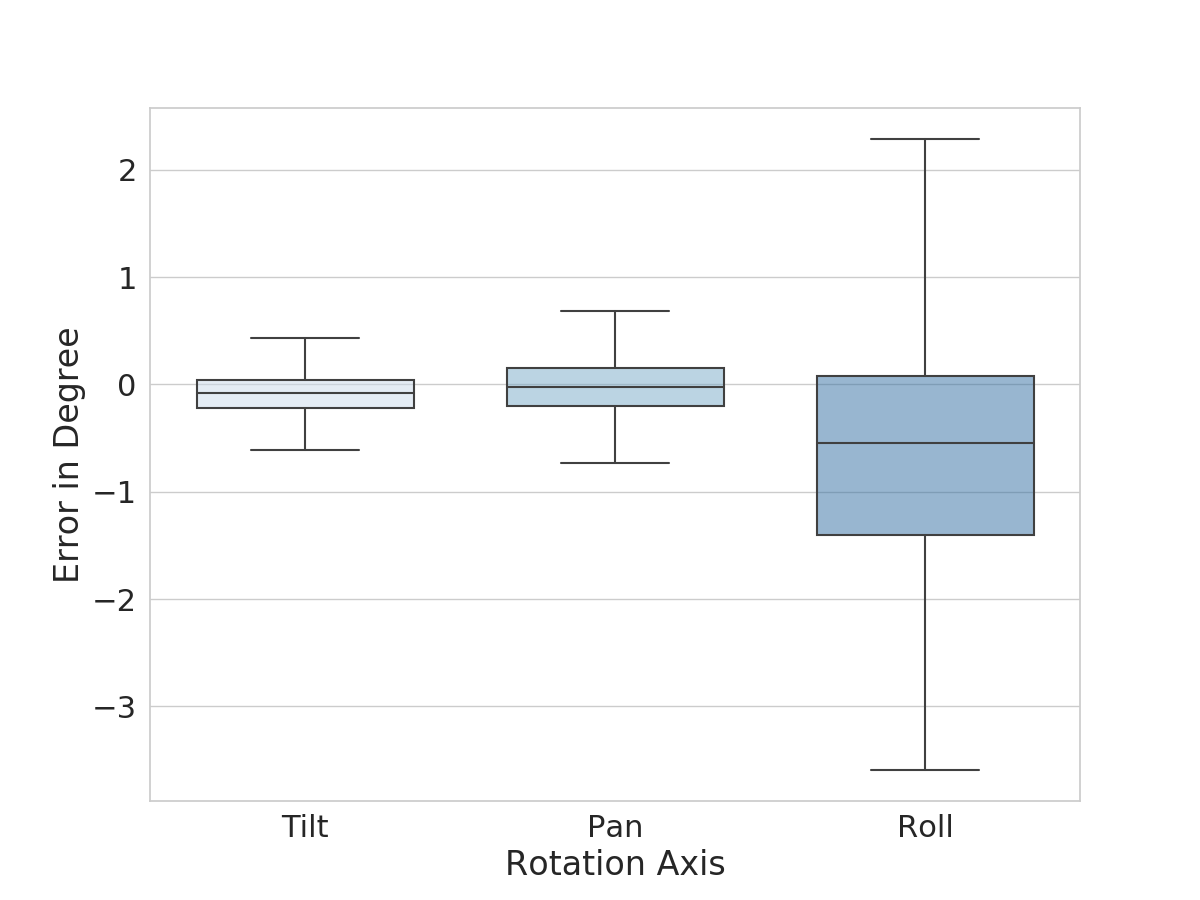}
        \caption{Initial errors: tilt = \SI{7.46}{\degree}, pan = \SI{-7.99}{\degree}, roll = \SI{-2.31}{\degree}}
    \end{subfigure}%
    \caption{Resulting error distributions over all $\mathcal{T}_1$ test set samples for two different static decalibrations. The means are close to zero, which shows that a temporal averaging could further improve the performance of our approach.}
    \label{fig:error-distributions-static}
\end{figure*}

We also evaluated our approach by applying the same decalibration to all samples of the test set $\mathcal{T}_1$. This is a more realistic setting. In this manner we evaluated 100 different decalibrations. In particular, we computed the error for each decalibration as the mean error over all samples. This way our approach achieved on average decalibration errors of \SI{0.21}{\degree} for tilt, \SI{0.35}{\degree} for pan and \SI{1.33}{\degree} for roll. As shown in~\fig{fig:error-distributions-static-all}~\textbf{(b)}, taking the average over all sample errors with the same static decalibration significantly reduces the error variance compared to using only a single frame for calibration like in the random decalibration setting shown in~\fig{fig:error-distributions}~\textbf{(b)}. Our model is able to reduce the static errors over all samples towards a distribution with approximately zero mean, as shown for two examples in~\fig{fig:error-distributions-static}. This indicates that temporal averaging of the estimated decalibration corrections as proposed in~\sect{sec:refinement} could be a suitable method to further improve accuracy and robustness.

\subsubsection{Generalization}
\begin{figure*}[ht]
    \centering
    \begin{subfigure}[b]{0.33\textwidth}
        \includegraphics[trim=53 55 46 60,clip,width=0.988\textwidth]{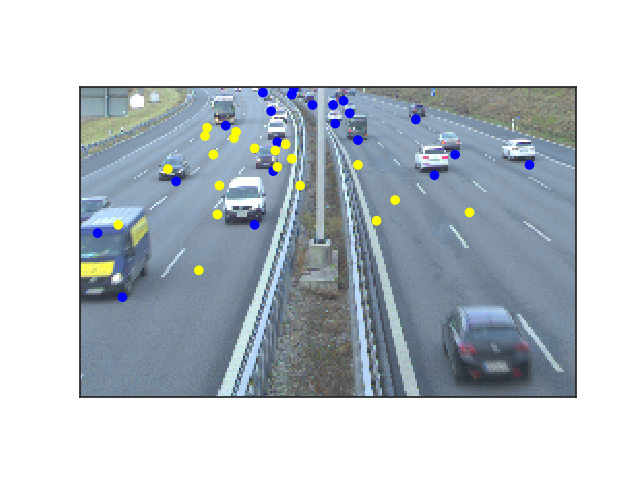}
        \caption{Initial}
    \end{subfigure}%
    \begin{subfigure}[b]{0.33\textwidth}
        \includegraphics[trim=53 55 46 60,clip,width=0.988\textwidth]{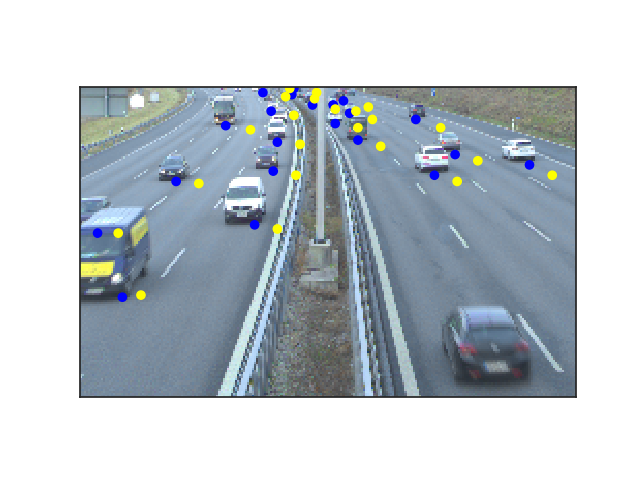}
        \caption{Coarse Network}
    \end{subfigure}%
    \begin{subfigure}[b]{0.33\textwidth}
        \includegraphics[trim=53 55 46 60,clip,width=0.988\textwidth]{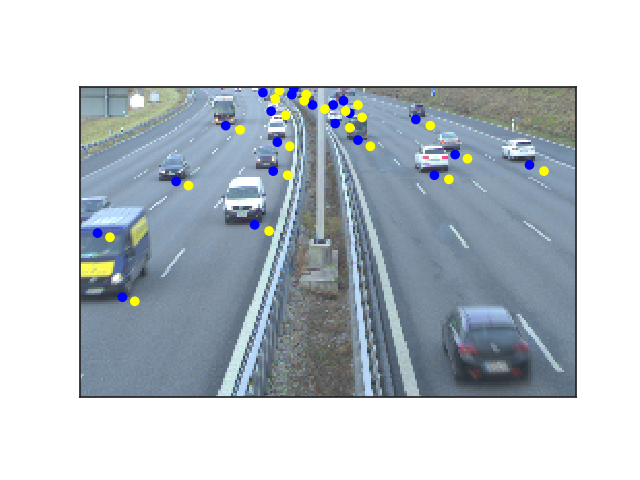}
        \caption{Fine Network}
    \end{subfigure}%
    
    \caption{Calibration results of applying our approach which was trained on one sensor setup to an unseen sensor setup (test set $\mathcal{T}_2$). Blue points represent the projected radar detections using the ground truth calibration and yellow projections using the calibration at (a) the initial stage, (b) after applying the coarse network and (c) after applying the fine network. Even though our model has never observed this perspective during training and the true calibration differs from the one in the training data, it generalizes and achieves reasonable results.}
    \label{fig:qualitative-mp2}
\end{figure*}

To demonstrate the generalization capability of our approach we applied it to test set $\mathcal{T}_2$, which is obtained from a sensor setup located at a different gantry bridge that was not included in the training data and has never been observed before. In this case the trajectory of the street is different and thus the distribution of vehicles in the image. Besides, the true extrinsic calibration differs from the first sensor setup and the perspective of the camera observing the vehicles changed. Despite these challenges, our approach achieved reasonable results for random decalibrations with average errors of \SI{0.36}{\degree} for tilt, \SI{1.88}{\degree} for pan and \SI{2.83}{\degree} for roll~(\fig{fig:qualitative-mp2}). While the performance dropped compared to the sensor setup used for training, it indicates that our approach is able to generalize if trained on a more diverse dataset with different perspectives and road segments. Furthermore, the achieved results already vastly reduce manual calibration efforts. It can also support other calibration methods in practice, as the difficulty to match correspondences is greatly reduced.

\section{Conclusion}
The manual calibration of sensors in an ITS is tedious and expensive, especially concerning sensor orientations. For radars and cameras there is a lack of automatic calibration methods due to the sparsity and absence of descriptive features in radar detections. We addressed this problem and presented the first approach for automatic rotational calibration of radar and camera sensors without the need of dedicated calibration targets. Our approach consists of two convolutional neural networks that are trained with a boosting-inspired learning regime. We evaluated our method on a real-world dataset that we recorded on a German highway. Our method achieves precise rotational calibration of the sensors and is robust to missing vehicle detections, multiple detections for single vehicles and noise. We demonstrated its generalization capability and achieved reasonable results by applying it on a second measurement point with a different viewing angle on the highway and vehicles. This drastically reduces the efforts of manual calibration.

We expect that in the future the generalization capabilities of our approach could be further improved by using a more diverse dataset that includes multiple camera perspectives. Furthermore, as sensors record a time series, sequences of frames could be used for iterative calibration with a recurrent neural network to increase calibration precision and robustness. As after the application of our approach the association of radar detections with vehicle detections in the image can be easily achieved with nearest-neighbor algorithms, the final results could be revised by solving a classic, convex optimization problem.

\addtolength{\textheight}{-0cm}   % This command serves to balance the column lengths

%%%%%%%%%%%%%%%%%%%%%%%%%%%%%%%%%%%%%%%%%%%%%%%%%%%%%%%%%%%%%%%%%%%%%%%%%%%%%%%%

\bibliographystyle{IEEEtran}
\bibliography{IEEEabrv,include/references}

\begin{thebibliography}{10}
\providecommand{\url}[1]{#1}
\csname url@rmstyle\endcsname
\providecommand{\newblock}{\relax}
\providecommand{\bibinfo}[2]{#2}
\providecommand\BIBentrySTDinterwordspacing{\spaceskip=0pt\relax}
\providecommand\BIBentryALTinterwordstretchfactor{4}
\providecommand\BIBentryALTinterwordspacing{\spaceskip=\fontdimen2\font plus
\BIBentryALTinterwordstretchfactor\fontdimen3\font minus
  \fontdimen4\font\relax}
\providecommand\BIBforeignlanguage[2]{{%
\expandafter\ifx\csname l@#1\endcsname\relax
\typeout{** WARNING: IEEEtran.bst: No hyphenation pattern has been}%
\typeout{** loaded for the language `#1'. Using the pattern for}%
\typeout{** the default language instead.}%
\else
\language=\csname l@#1\endcsname
\fi
#2}}

\bibitem{Wang2017}
M.~Wang, L.~Jiang, W.~Lu, and Q.~Ma, ``Detection and tracking of vehicles based
  on video and {2D} radar information,'' \emph{International Conference on
  Intelligent Transportation (ICIT)}, 2017.

\bibitem{Roy2009}
A.~Roy, N.~Gale, and L.~Hong, ``Fusion of doppler radar and video information
  for automated traffic surveillance,'' \emph{International Conference on
  Information Fusion (FUSION)}, 2009.

\bibitem{Hinz2017}
G.~Hinz, M.~Buechel, F.~Diehl, G.~Chen, A.~Kraemmer, J.~Kuhn,
  V.~Lakshminarasimhan, M.~Schellmann, U.~Baumgarten, and A.~Knoll, ``Designing
  a far-reaching view for highway traffic scenarios with {5G}-based intelligent
  infrastructure,'' \emph{8. Tagung Fahrerassistenz}, 2017.

\bibitem{Zhang2004}
Q.~Zhang and R.~Pless, ``Extrinsic calibration of a camera and laser range
  finder (improves camera calibration),'' \emph{International Conference on
  Intelligent Robots and Systems (IROS)}, 2004.

\bibitem{Zhi2007}
G.~Zhi, Z.~Sidong, Z.~Wei, and Z.~Yunyi, ``A high-precision calibration
  technique for laser measurement instrument and stereo vision sensors,''
  \emph{International Conference on Electronic Measurement and Instruments
  (ICEMI)}, 2007.

\bibitem{Scaramuzza2007}
D.~Scaramuzza, A.~Harati, and R.~Siegwart, ``Extrinsic self calibration of a
  camera and a {3D} laser range finder from natural scenes,''
  \emph{International Conference on Intelligent Robots and Systems (IROS)},
  2007.

\bibitem{Chien2016}
H.-J. Chien, R.~Klette, N.~Schneider, and U.~Franke, ``Visual odometry driven
  online calibration for monocular lidar-camera systems,'' \emph{International
  Conference on Pattern Recognition (ICPR)}, 2016.

\bibitem{Levinson2013}
J.~Levinson and S.~Thrun, ``Automatic online calibration of cameras and
  lasers,'' \emph{Robotics: Science and Systems (RSS)}, 2013.

\bibitem{Kendall2015}
A.~Kendall, M.~K. Grimes, and R.~Cipolla, ``{PoseNet}: A convolutional network
  for real-time {6-DOF} camera relocalization,'' \emph{International Conference
  on Computer Vision (ICCV)}, 2015.

\bibitem{RegNet}
N.~Schneider, F.~Piewak, C.~Stiller, and U.~Franke, ``{RegNet}: Multimodal
  sensor registration using deep neural networks,'' \emph{Intelligent Vehicles
  Symposium (IV)}, 2017.

\bibitem{Liu2018}
H.~{Liu}, Y.~{Liu}, X.~{Gu}, Y.~{Wu}, F.~{Qu}, and L.~{Huang}, ``A
  deep-learning based multi-modality sensor calibration method for usv,''
  \emph{International Conference on Multimedia Big Data (BigMM)}, 2018.

\bibitem{CalibNet}
G.~{Iyer}, R.~K. {Ram}, J.~K. {Murthy}, and K.~M. {Krishna}, ``{CalibNet}:
  Self-supervised extrinsic calibration using {3D} spatial transformer
  networks,'' \emph{International Conference on Intelligent Robots and Systems
  (IROS)}, 2018.

\bibitem{Helmick1993}
R.~E. Helmick and T.~R. Rice, ``Removal of alignment errors in an integrated
  system of two {3-D} sensors,'' \emph{Transactions on Aerospace and Electronic
  Systems (T-AES)}, 1993.

\bibitem{Li2006}
Z.~Li and H.~Leung, ``An expectation maximization based simultaneous
  registration and fusion algorithm for radar networks,'' \emph{Canadian
  Conference on Electrical and Computer Engineering (CCECE)}, 2006.

\bibitem{Natour2015}
G.~E. Natour, O.~A. Aider, R.~Rouveure, F.~Berry, and P.~Faure, ``Radar and
  vision sensors calibration for outdoor {3D} reconstruction,''
  \emph{International Conference on Robotics and Automation (ICRA)}, 2015.

\bibitem{Persic2017}
J.~Per{\v{s}}i{\'{c}}, I.~Markovi{\'{c}}, and I.~Petrovi{\'{c}}, ``Extrinsic
  {6DoF} calibration of {3D} lidar and radar,'' \emph{European Conference on
  Mobile Robots (ECMR)}, 2017.

\bibitem{Song2017}
C.~Song, G.~Son, H.~Kim, D.~Gu, J.~H. Lee, and Y.~Kim, ``A novel method of
  spatial calibration for camera and {2D} radar based on registration,''
  \emph{International Congress on Advanced Applied Informatics (AAI)}, 2017.

\bibitem{Howard2017}
A.~G. Howard, M.~Zhu, B.~Chen, D.~Kalenichenko, W.~Wang, T.~Weyand,
  M.~Andreetto, and H.~Adam, ``{MobileNets}: Efficient convolutional neural
  networks for mobile vision applications,'' \emph{arXiv:1704.04861 [cs.CV]},
  2017.

\bibitem{JiaDeng2009}
J.~Deng, W.~Dong, R.~Socher, L.-J. Li, K.~Li, and L.~Fei-Fei, ``{ImageNet}: A
  large-scale hierarchical image database,'' \emph{Conference on Computer
  Vision and Pattern Recognition (CVPR)}, 2009.

\bibitem{Lin2013}
M.~Lin, Q.~Chen, and S.~Yan, ``Network in network,'' \emph{International
  Conference on Learning Representations (ICLR)}, 2013.

\bibitem{Dropout2014}
N.~Srivastava, G.~Hinton, A.~Krizhevsky, I.~Sutskever, and R.~Salakhutdinov,
  ``Dropout: A simple way to prevent neural networks from overfitting,''
  \emph{Journal of Machine Learning Research (JMLR)}, 2014.

\bibitem{PReLu2015}
K.~He, X.~Zhang, S.~Ren, and J.~Sun, ``Delving deep into rectifiers: Surpassing
  human-level performance on {ImageNet} classification,'' \emph{International
  Conference on Computer Vision (ICCV)}, 2015.

\bibitem{Huynh2009}
D.~Q. Huynh, ``Metrics for {3D} rotations: Comparison and analysis,''
  \emph{Journal of Mathematical Imaging and Vision}, 2009.

\bibitem{Kuffner2004}
J.~J. {Kuffner}, ``Effective sampling and distance metrics for {3D} rigid body
  path planning,'' \emph{International Conference on Robotics and Automation
  (ICRA)}, 2004.

\bibitem{KingmaB14}
D.~P. Kingma and J.~Ba, ``Adam: {A} method for stochastic optimization,''
  \emph{International Conference on Learning Representations (ICLR)}, 2014.

\bibitem{Saxe2014}
A.~M. Saxe, J.~L. McClelland, and S.~Ganguli, ``Exact solutions to the
  nonlinear dynamics of learning in deep linear neural networks,''
  \emph{International Conference on Learning Representations (ICRA)}, 2014.

\bibitem{friedman2002stochastic}
J.~H. Friedman, ``Stochastic gradient boosting,'' \emph{Computational
  statistics and Data Analysis (CSDA)}, 2002.

\bibitem{Kanhere2010}
N.~K. Kanhere and S.~T. Birchfield, ``A taxonomy and analysis of camera
  calibration methods for traffic monitoring applications,'' \emph{Transactions
  on Intelligent Transportation Systems (T-ITS)}, 2010.

\end{thebibliography}

\end{document}